\documentclass[10pt,journal,compsoc]{IEEEtran}

\usepackage[nocompress]{cite}
\usepackage[caption=false,font=footnotesize,labelfont=sf,textfont=sf]{subfig}
\usepackage{ragged2e}
\usepackage{caption}
\usepackage{multirow}
\usepackage{array}
\usepackage[hidelinks]{hyperref}
\newcolumntype{C}[1]{>{\centering\arraybackslash}m{#1}}
\usepackage{graphicx}
\graphicspath{{./Images/}}

\begin{document}

\markboth{IEEE TRANSACTIONS ON PATTERN ANALYSIS AND MACHINE INTELLIGENCE,~Vol.~, No.~, Month~Year}
		{Shell \MakeLowercase{\textit{et al.}}: PATTERN ANALYSIS AND MACHINE INTELLIGENCE}


\title{SwiftFace: Real-Time Face Detection}

\author{
	Leonardo Ramos,
        Bernardo Morales
        \IEEEcompsocitemizethanks{
		\IEEEcompsocthanksitem L. Ramos and B. Morales are with Simón Bolívar University, Valle de Sartenejas, Caracas, Venezuela.
}
}

\IEEEtitleabstractindextext{
	\justify
	\begin{abstract}
	 Computer vision is a field of artificial intelligence that trains computers to interpret the visual world in a way similar to that of humans. Due to the rapid advancements in technology and the increasing availability of sufficiently large training datasets, the topics within computer vision have experienced a steep growth in the last decade. Among them, one of the most promising fields is face detection. Being used daily in a wide variety of fields; from mobile apps and augmented reality for entertainment purposes, to social studies and security cameras; designing high-performance models for face detection is crucial. On top of that, with the aforementioned growth in face detection technologies, precision and accuracy are no longer the only relevant factors: for real-time face detection, speed of detection is essential. SwiftFace is a novel deep learning model created solely to be a fast face detection model. By focusing only on detecting faces, SwiftFace performs 30\% faster than current state-of-the-art face detection models. Code available at \url{https://github.com/leo7r/swiftface}.
	\end{abstract}
	
	\begin{IEEEkeywords}
	Face detection, unsupervised deep learning, object detection, neural networks.
	\end{IEEEkeywords}
}

\maketitle


\IEEEraisesectionheading{\section{INTRODUCTION}\label{sec:introduction}}

\IEEEPARstart{A}{rtificial} intelligence, human attempts to make computers think as we do, has experienced noticeable growth in the past few years. However, these models have been around since the 1940s \cite{Walczak2019}.\\
\indent One important feature of the human brain is its ability to detect faces at a glance by quickly processing images perceived by the eyes. Based on the behavior of the brain’s neurons, artificial neural networks were born and, as early as the 1960s were being used for facial detection when Bledsoe, Wolf, and Bisson started using computers to identify the characteristic features of the human face \cite{Andreopoulos2013}. Following Bledsoe, Wolf, and Bisson’s work, in the 1970s, Goldstein, Harmon, and Lesk worked on an even more extensive and specific list of 22 features to be used as markers to detect and then identify human faces from a pool of photographs \cite{Goldstein1971}. A few decades later, Turk and Pentland developed a “near-real-time computer system” able to notice, locate and track a subject’s head movement by detecting characteristic features of human's face \cite{Turk1991}.

\begin{figure}[!htb]
\centering
\includegraphics[width=3.5in,height=2.5in]{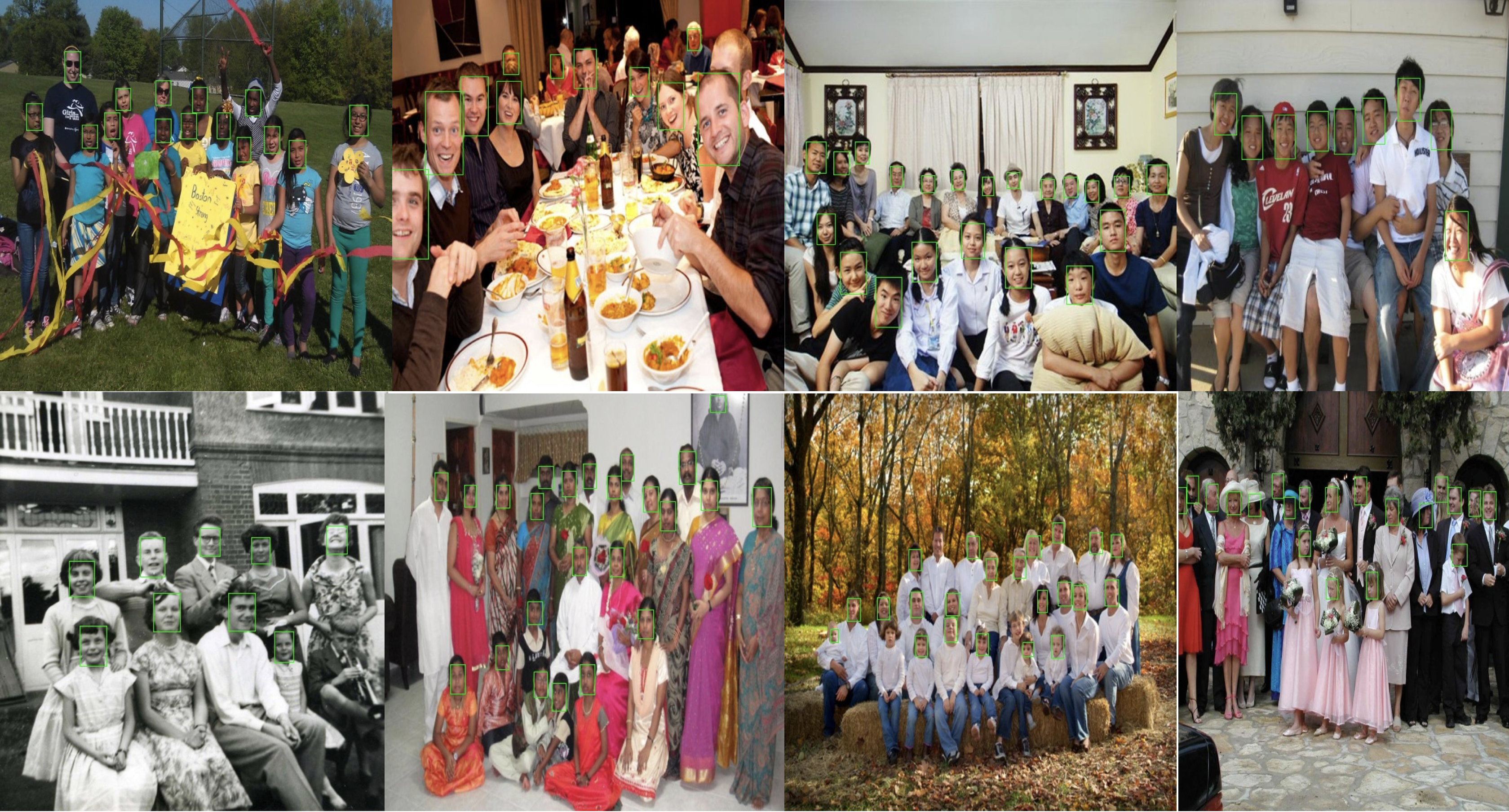}
\captionsetup{justification=centering,margin=0.3in}
\caption{Different face detection samples.}
\label{fig_1}
\end{figure}

\indent These early advances on facial detection were hindered by the limitations of the technology of that time but served as the basis for future research in the area. Hence, as the internet grew through the early 2000s and made available even bigger sets of images that could be used for analysis and training, facial detection models flourished rapidly, making it possible for machines to accurately and quickly locate faces in both photos and videos. Now, a couple of decades later, we have cellphones with integrated cameras, social media that encourage users to take and share numerous photos and videos, and incredibly powerful computer devices have become easily accessible for a great number of people \cite{Anil2016}.\\
\indent With the growth of resources available to make experiments in facial detection, so has grown the number of fields on which its application means an increase in productivity and/or ease of use for different practices in daily living \cite{Kumar2018}. These developments in technology mean that precision and accuracy are no longer the only relevant characteristics when choosing an image detection model to implement. A model’s capacity to detect faces quickly is necessary to accommodate the day-to-day usage of different practices. A detection algorithm needs then not only to be accurate but to be able to respond fast in order to successfully play its role in systems such as face-identification log-in features, social networks image tagging, photography and entertainment apps filters, augmented reality devices, surveillance and security, market research, among others.\\
\indent Top-performing real-time face detection models have achieved performance rates of over 99\% in 2020, compared to the 96\% for leading algorithms in 2014, which have allowed for facial detection models to be applied successfully even in emergencies \cite{Tikoo2017}. Let’s take, for example, our current situation with the COVID-19 pandemic. Face detection has reached such impressive performance rates that, even when people wear masks, facial detection can efficiently locate individual’s faces, making it possible to  track social interactions of potentially contagious individuals, allowing for an almost immediate response, hence minimizing the impact of the virus \cite{findfacepro2020}.\\
\indent The general trend in computer vision is to make deeper artificial networks, with numerous blocks and layers to achieve higher accuracy \cite{Szegedy2015}. However, those higher accuracy rates are paired to slower responses due to the heavier computational cost. These accurate but slow face detection models might fall short when facing real-world applications, which require real-time performance in equipment with limited computational capacity. \\
\indent In section two of this article some of the current state-of-the art image detection models are presented, with focus in performance comparisions, specifically those related to the model’s accuracy, recall, and speed.\\
\indent In section three, SwiftFace is presented. A faster approach to face detection based on previous image detection models thought to maintain accuracy while improving the model’s speed.\\
\indent In section four, SwiftFace’s performance is brought into comparison with two image detection models, widely used due to their high accuracy rates and out-standing speed even when used in low-end devices.\\
\indent In section five, possible real applications of SwiftFace are presented, with a focus on SwiftFace’s strong point: top-performing speed without loss of accuracy.\\
\indent In section six, the concluding remarks from this article are presented and related future research topics are proposed.


\section{RELATED WORK}
With the growth of available computing power and images for training and testing data, the number of works in developing object detection algorithms has risen. 
Face detection, as a special case of object detection, is usually improved based on algorithms originally meant to be used in multi-class object detection. These algorithms are usually based on deep learning and artificial neural networks and, as such, can handle and process large amounts of data in a relatively short time \cite{Al-allaf2014}.\\
\indent Convolutional Neural Networks (CNNs) is one of the main architectures used in computer vision. As opposed to Multi-Layer Perceptrons, CNN has filters that in turn generate convoluted layers from which information is extracted, instead of just fully-connected layers. The use of a convolution layer allows for the model to extract relational information and identify patterns drawn from an input. Also, since the layers are not fully-connected, CNN has fewer values that need to be learned and updated than a regular MLP neural network, as filters perform much better in image recognition problems.\\
\indent Based on Convolutional Neural Networks, some of the most used image recognition models are presented in the following subsections. 

	\begin{figure}[!htb]
	\centering
	\includegraphics[width=3.5in]{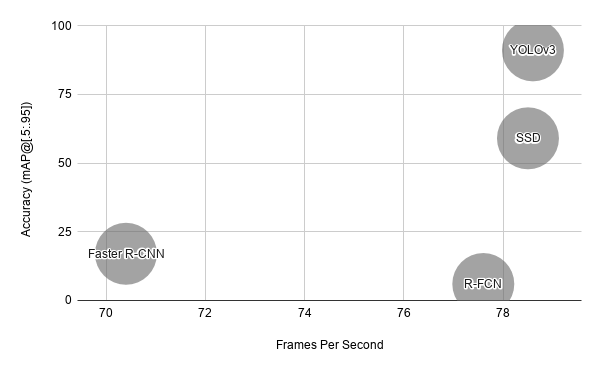}
	\captionsetup{justification=centering,margin=0.3in}
	\caption{Graphic comparison between the main performance indicators of current state-of-the-art image detection models.}
	\label{fig_2}
	\end{figure}

	\subsection{YOLO}
	You Only Look Once (YOLO) \cite{Redmon2016} is an algorithm based on CNN designed to operate with fast response speeds as an object detector in production systems. It was developed as a one-step process involving detection and classification. YOLO uses a single-step neural net-work to predict class probabilities from the input images directly in one evaluation \cite{Bochovsky2020}. The CNN uses several filters to divide the input image into weighted grid cells and then predicts the and classification values for the image by studying batches of cells and their relationship with each other. At 78.6 FPS, YOLO’s speed is several times bigger than the most common two-stage detectors, and the model’s accuracy at mAP of 91 looks relatively high when compared with direct counterparts \cite{Chen2020}.\\
	\indent Since the first YOLO model came out, several variations have been developed; the Tiny YOLO version being one of the fastest among them. The Tiny-YOLO architecture is around 400\% faster than the original version, being able to achieve 244 FPS on a computer with a single GPU \cite{Bonn2020}.

	\subsection{RCNN}
	Regional-based Convolutional Neural Networks (R-CNN) is also a CNN based algorithm used for image detection \cite{Zang}. R-CNN uses selective search to extract certain regions from the target image. Therefore, the problem is reduced from the full number of cells in which the image was originally divided, to just regions that group some adjacent cells. From each region proposal, a feature vector is extracted and fed into a regular CNN and then evaluated \cite{Jiang2017}. R-CNN results in high accuracy, but even reducing the image analysis according to the regions, this model doesn’t achieve acceptable real-time performance, which makes it unviable for real-time applications.\\
	\indent To correct these drawbacks of R-CNN, several variations have been developed in order to obtain similar object detection algorithms but with higher speeds. Fast R-CNN and Faster R-CNN are two of the most notables. As the name suggests, Faster R-CNN has the best performance in terms of speed, with 70.4 FPS, but with a lower accuracy at  mAP 17. \\

	\subsection{SSD}
	Single Shot Detection (SSD) is a feed-forward CNN based architecture that produces a fixed-size group of bounding boxes and classification values for the presence of object class instances within those boxes \cite{Liu2016}. After the boxing analysis, a non-maximum suppression step is applied in order to generate the output values. The network layers are based on a standard CNN architecture used for high-quality image classification, where convolution filters for each cell are used to make the predictions \cite{Li2018}.\\
	\indent SSD has remarkable performance indicators at both speed (78.5 FPS) and accuracy (mAP 59).\\

	\subsection{R-FCN}
	Region-based Fully Convolutional Networks (R-FCN) was devised as an accurate and efficient object detection model \cite{Dai2016}. R-FCN addresses the computational cost of region-based networks and proposes a solution by using position-sensitive score maps, or regional feature maps where each detects and scores their corresponding region of the image \cite{Tang2020}. With this, combining the scores results in the image accurately being located.\\
	\indent By reducing the computations needed compared to other regional-based convolutional neural networks, R-FCN tends to perform faster than its counterparts at 77.6 FPS, but has an even lower accuracy given by a mAP of 6.\\

	\begin{table}[!htb]
	\renewcommand{\arraystretch}{2.5}
	\captionsetup{justification=centering,margin=0.3in}
	\caption{Main Performance Indicators of Current State-of-the-Art Image Detection Models}
	\label{table_1}
	\centering
	\begin{tabular}{p{3.5cm} C{2cm} C{2cm}}
	\hline
	Model & Accuracy & FPS\\
	\hline
	Faster R-CNN & 70.4 & 17\\
	R-FCN & 77.6 & 6\\
	SSD & 78.5 & 59\\
	YOLOv3 & 78.6 & 91\\
	\hline
	\end{tabular}
	\end{table}


\section{SWIFTFACE, A FASTER APPROACH}
SwiftFace is an architecture based on the tinyYOLO model, specially devised to focus on face detection, maintaining its accuracy while improving its speed. SwiftFace was trained using the WIDERFACE \cite{yang2016} dataset, making improvements to the original Tiny-YOLO architecture and tuning-up the model to improve greatly the detection speed and still achieve similar accuracy rates when detecting faces.

\begin{figure}[!htb]
\centering
\includegraphics[width=3.5in,height=5.5in]{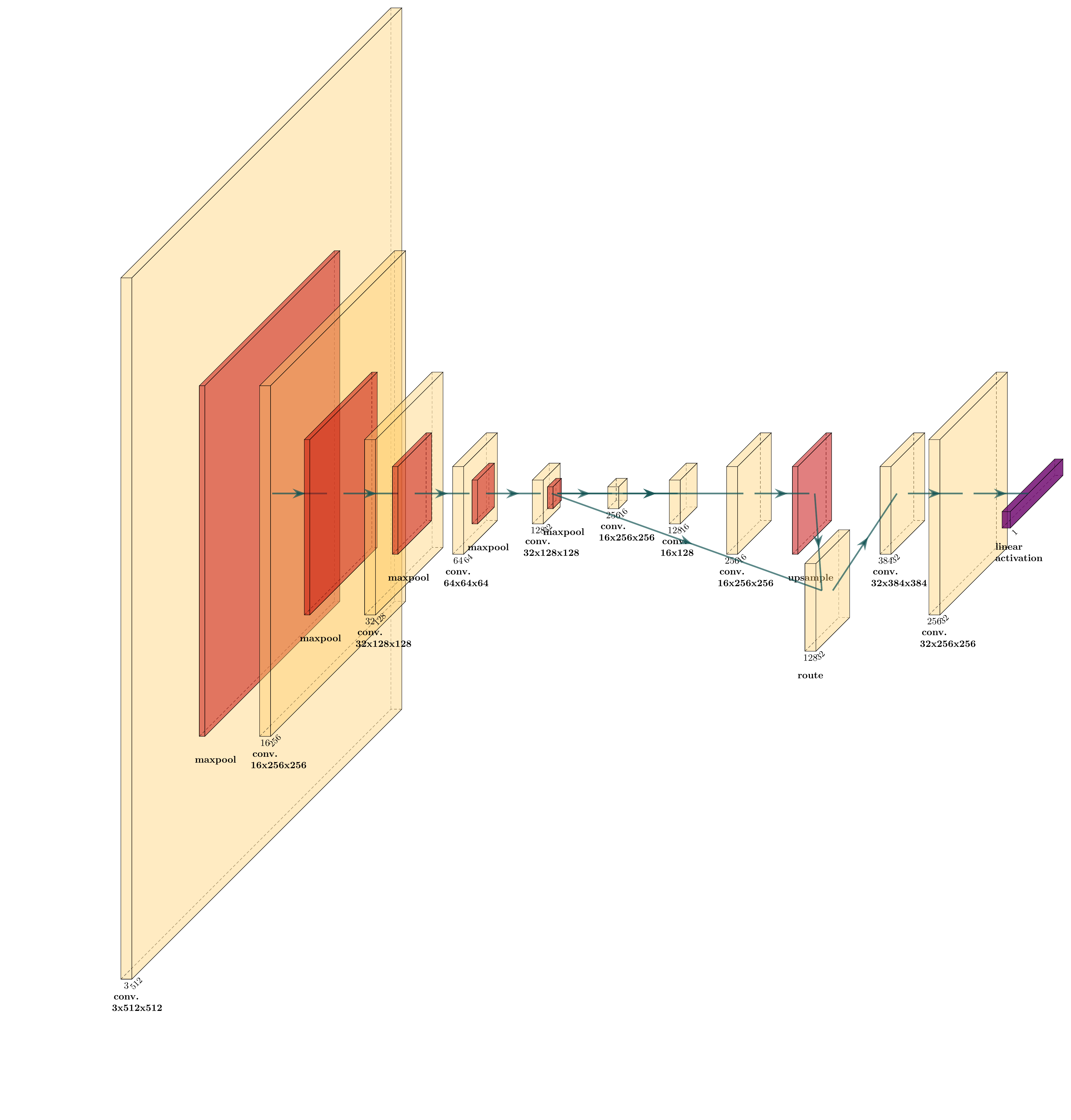}
\captionsetup{justification=centering,margin=0.3in}
\caption{A visual description of SwiftFace’s architecture.}
\label{fig_3}
\end{figure}

\begin{table}[!htb]
\renewcommand{\arraystretch}{2.5}
\captionsetup{justification=centering}
\caption{SwiftFace layers, including the number of filters, the filter’s size and stride and the layer’s input and output dimensions.}
\label{table_1}
\centering
\begin{tabular}{ccccc}
\hline
Layer & Filters & Size/Stride & Input & Output\\
\hline
0 conv & 16 & 3x3 / 1 &512x512x3 & 512x512x16\\
1 max & & 2x2 / 2 &512x512x16 &256x256x16\\
2 conv & 32 & 3x3 / 1 &256x256x16 & 256x256x32\\
3 max & & 2x2 / 2 &256x256x32 & 128x128x32\\
4 conv & 64 & 3x3 / 1 & 128x128x32 &  128x128x64\\
5 max & & 2x2 / 2 & 128x128x64 & 64x64x128\\
6 conv & 128 & 3x3 / 1 &64x64x64 & 64x64x64\\
7 max & & 2x2 / 2 &64x64x128 & 32x32x128\\
8 conv & 256 & 3x3 / 1 &32x32x128 & 32x32x256\\
9 max & & 2x2 / 2 &32x32x256 & 16x16x256\\
10 conv & 128 & 3x3 / 1 &16x16x256 & 16x16x512\\
11 conv & 18 & 1x1 / 1 &16x16x512 & 16x16x18\\
12 yolo & \multicolumn{4}{p{6cm}}{mask = 3,4,5; anchors = 10,14,23,27,37,58,81,82,135,169,344,319 classes = 1; num = 6; jitter = 0.3; ignorethresh = 0.7; truththresh = 1; random = 1} \\
13 route & 9 &  &  & 16x16x256\\
14 conv & 128 & 1x1 / 1 &16x16x256 & 16x16x128\\
15 upsample & & 2x / 1 &16x16x128 & 32x32x128\\
16 route & 158 & & & 32x32x384\\
17 conv & 256 & 3x3 / 1 &32x32x384 & 32x32x256\\
18 conv & 18 & 1x1 / 1 &32x32x256 & 32x32x18\\
\hline
\end{tabular}
\end{table}

\indent Swiftace is made by 18 layers, as opposed to Tiny-YOLO’s  23. The first ten layers are alternating pairs of convolution layers and pools, then there’s an additional convolution layer right before one that runs the same as the original Tiny-YOLO algorithm. Then a routing layer concatenates with the output of a previously defined layer. The upsample layer doubles the dimension of the input right before another route layer. Finally, two additional convolution layers extract information from their respective inputs and sending the data towards the final YOLO layer which then delivers the corresponding output.\\
\indent Being CNN based, SwiftFace's core feature is its convolutional layers. Their role is to extract features from the input image while preserving the information provided by the dimensional relationship between the cells of that image.\\
\indent In order to obtain the best results from the convolution neural network, specific parameters were set based on modifications on the YOLO original structure:\\
\indent The total amount of classes was set to one since the goal of SwifFace is face detection only. The maximum batches size parameter was established following the recommendation that it should be larger than the number of training images (~13.000), hence, it was set at 15.000. Steps then were set in the range of 80\%~90\% of the maximum batches, which meant between 12.000 and 13.500. The input image size was 512x512. The last layer function was set to be linear activation since only one class is expected as an output.\\
\indent SwiftFace was then trained and re-trained several times using the WIDERFACE dataset in order to come up with the best performant model for the face detection task. Paired with that, since SwiftFace focuses only on one class, as opposed to YOLO’s 80 classes, it performs faster than its counterpart. 


\section{BENCHMARKS}
We tested our SwiftFace model against two of its direct competitors: YOLOv4 and Tiny-YOLOv4.\\

\begin{figure}[!htb]
\centering
\includegraphics[width=3.5in,height=2.5in]{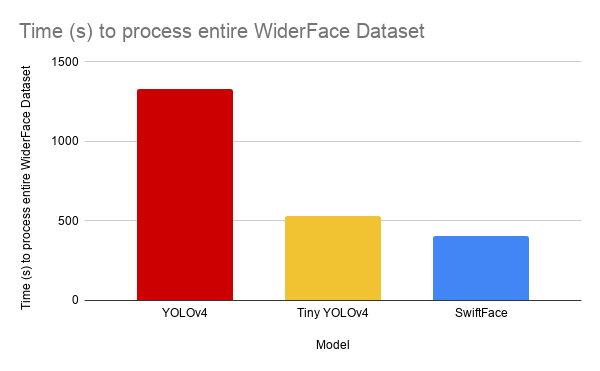}
\captionsetup{justification=centering,margin=0.3in,belowskip=15pt}
\caption{Speed comparison against YOLOv4 and TinyYOLOv4.\\}
\label{fig_4}
\end{figure}

\indent In terms of speed, YOLOv4 took 1329 seconds to process the entire WIDERFACE dataset, for a FPS 0f 12.1; while SwiftFace managed to do it in just 470 seconds, achieving a FPS of 39.5, making our model 69\% faster than the widely used YOLOv4.  When testing Tiny-YOLOv4, a model designed specifically to yield high-performance rates in terms of speed, the time spent to process the entire WIDERFACE dataset was 533 seconds, yielding a FPS of 30.1, which makes it 24\% slower than SwiftFace. Figure 4 shows a graphic comparison between the processing speed of all three tested models.\\

\begin{table}[!htb]
\renewcommand{\arraystretch}{2.5}
\captionsetup{justification=centering,margin=0.3in}
\caption{Time to Process the Entire WIDERFACE Dataset}
\label{table_1}
\centering
\begin{tabular}{p{3.5cm} C{2cm} C{2cm}}
\hline
Model & Time (s) & Number of images\\
\hline
YOLOv4 & 1329 & 16067\\
Tiny YOLOv4 & 533 & 16067\\
SwiftFace & 407 & 16067\\
\hline
\end{tabular}
\end{table}

\indent In order to compare the models in terms of accuracy, we calculated their mean average precision (mAP). For its mAP (.5:.95), Tiny-YOLOv4 had a mAP of 54\%, versus SwiftFace’s 51\%. Despite the difference in speed performance, accuracy-wise both models perform quite similarly, which puts SwiftFace in the front line of models that could be used in different real-time applications without falling behind top-tiers object detection algorithms.

\begin{figure}[!htb]
\centering
\includegraphics[width=3.5in]{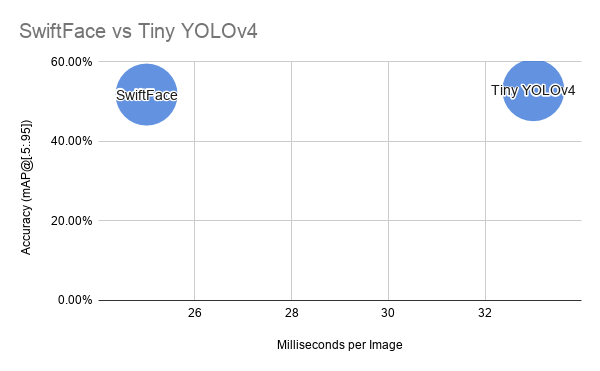}
\captionsetup{justification=centering,margin=0.3in}
\caption{Speed comparison against YOLOv4 and TinyYOLOv4.}
\label{fig_5}
\end{figure}


\section{REAL-LIFE APPLICATIONS}
The range of applications for software such as SwiftFace is wide; and it’s getting wider every year thanks to technological advances. Being a fast-performing face detection algorithm that competes in terms of accuracy with top-of-the-line image detection models, SwiftFace provides state-of-the-art performance for mobile devices, low-end devices, and edge computing. It can be used in real-time applications and efficiently reduce the amount of human input in different fields. Applications involving low-end devices, which are more affordable and easily available, especially benefit from algorithms like SwiftFace, since it reduces the hardware and deployment costs of face detection applications in production. Among the most promising uses, we find: 

	\subsection{Human-Machine interaction}
	Human-computer interaction systems are evolving to be more independent and to need less and less human input. One way of achieving this is making human-machine interactions control schemes dependent on intuitive human features, such as the face. Cameras that automatically take a picture when 	detecting a smiling face are an example of this.
	
	\subsection{Mobile apps and entertainment}
	Talking a bit deeper about cameras, some cellphone cameras use face detection for autofocus. This type of face detection is also useful for selecting regions of interest in photo slideshows, automatic face-tagging in photo-graphs is social media, face-priority algorithms when displaying images previews in profile pictures and in-ternet posts, etcetera.
	
	\subsection{Work from Home and online education}
	With the current worldwide-pandemic situation, online solution to daily activities like work and education have been rapidly growing.  Fast-performing and accurate face detection models work as a as well as the main step in relevant processes such as quick and easy attendance-check, face-focusing video and automatic background changing.
	
	\subsection{Market research}
	Face detection will have an impact on marketing thanks to its application in market research. Instead of having a human performing the tedious task of recognizing and counting how many possible buyers’ glance at a product in a display, a face detection algorithm can do the same at a lower cost and possibly even with higher accuracy, by just integrating a camera to detect face that walks by. With further processing, additional information can be gathered from the public, such as age and gender, to build even more specific buyer personas.
	
	\subsection{Access control}
	A quick and efficient face detection phase in biometric access control, whether it is face identification to unlock mobile devices or a company’s biometric access to its office building, is essential for the proper performance of the access system. 


\section{CONCLUSION AND FUTURE RESEARCH}
In a world where face detection and face recognition are going to be a central part of the technological develop-ment, affecting in fields as diverse as business, entertainment, and security, SwiftFace is a lighter face detec-tion model that performs 30\% faster than state-of-the-art object detection models. This will help future research-ers, engineers, and entrepreneurs to build world-class applications to advance even more the growing world-wide technological development.\\ 
\indent Face recognition in low-end or edge devices is still an unsolved problem with the current models. SwiftFace could provide an affordable, easy way of bringing face detection to these devices. In addition, SwiftFace offers numerous possible contributions to the field of face recognition. With similar accuracy performance to top-of-the-line models, but with better response times, SwiftFace has great potential for real-time face detection phases in face recognition models. \\
\indent In general, faster object detection algorithms tend to have lower accuracy rates than their slower counterparts. Nevertheless, with the impressive rise in the accuracy of computer vision models in the past years, this gap is sure to close with further improvements. As such, even SwiftFace could improve its rates compared to top-accuracy models like faster RCNN to provide better detections.\\
Future research should be carried in order to translate the results from SwiftFace to a face recognition model, to improve and compare inference time against state-of-the-art models such as RetinaNet, OpenFace, and FaceNet. 


\bibliography{SwiftFace}
\bibliographystyle{IEEEtran}

\end{document}